\documentclass[10pt, a4paper]{article}
\usepackage{lrec}
\usepackage{caption}
\usepackage{multibib}
\newcites{languageresource}{Language Resources}
\usepackage{graphicx}
\usepackage{tabularx}
\usepackage{soul}
\usepackage{algorithm,algorithmic}
\usepackage[T1]{fontenc}
\usepackage{textgreek}
\usepackage{multirow}
\usepackage{enumitem}
\usepackage{amsmath}
\usepackage{xcolor}
\setlist{noitemsep}
\usepackage{epstopdf}
\usepackage[latin1]{inputenc}

\usepackage{hyperref}
\usepackage{xstring}

\title{Preparing Bengali-English Code-Mixed Corpus for Sentiment Analysis of Indian Languages}

 \name{Soumil Mandal\textsuperscript{1}, Sainik Kumar Mahata\textsuperscript{2}, Dipankar Das\textsuperscript{3}}

 \address{\\
          \textsuperscript{1}Department of Computer Science \& Engineering, SRM University, Chennai \\\textsuperscript{2,3}Department of Computer Science \& Engineering, Jadavpur University, Kolkata \\
          \textcolor{black!50}{\{soumil.mandal, sainik.mahata, dipankar.dipnil2005\}@gmail.com}\\}

\abstract{
Analysis of informative contents and sentiments of social users has been attempted quite intensively in the recent past. Most of the systems are usable only for monolingual data and fails or gives poor results when used on data with code-mixing property. To gather attention and encourage researchers to work on this crisis, we prepared gold standard Bengali-English code-mixed data with language and polarity tag for sentiment analysis purposes. In this paper, we discuss the systems we prepared to collect and filter raw Twitter data. In order to reduce manual work while annotation, hybrid systems combining rule based and supervised models were developed for both language and sentiment tagging. The final corpus was annotated by a group of annotators following a few guidelines. The gold standard corpus thus obtained has impressive inter-annotator agreement obtained in terms of Kappa values. Various metrics like Code-Mixed Index (CMI), Code-Mixed Factor (CF) along with various aspects (language and emotion) also qualitatively polled the code-mixed and sentiment properties of the corpus.  
 \\ \newline  }

\begin{document}

\maketitleabstract
\linespread{1}
%\lipsum[1-4]
\section{Introduction}

India has a linguistically diverse and vast diaspora due to its long history of contact with foreigners. English, one of those borrowed languages, became an integral part of the Indian education system and has been recognized as one of the official languages as well, thus giving rise to a population where bilingualism is very common. This kind of language diversity coupled with various dialects instigates frequent code-mixing in India. This phenomenon has become even more transparent with the rise of social networking sites like Twitter and Facebook and also instant messaging services like WhatsApp etc. The writing style in such media indicates phonetic typing transliterated in Roman, generally mixed with English words through code-mixing and also Anglicism. Three facts are involved in this sort of code-mixing cases, 1. lack of knowledge in using appropriate native words, 2. typing convenience and 3. popularity of Roman script to cater to a large set of audience. 
\\ 
 \hspace*{0.5cm} Social networking services has been gaining popularity very rapidly since their first appearance and has led to an exponential growth of minable data which is rich and informative. In developing countries where majority of the population are bilinguals, in social media data, we frequent observe a unique trend in typing where two or more languages are mixed for expression known as code-mixing.  It is also observed that such code-mixed data are growing rapidly in WWW because multilingual users in social networks frequently share their sentiments and thus it becomes an important  task to mine and analyze such data for gathering crucial informatics related to sentiment too. However, the complexity involved in mixing of multiple rules of grammars, scripts, use of transliteration in such code-mixed data possesses a big challenge for NLP tasks. Thus, it becomes an ever so important task to solve this problem since a huge chunk of the data on social media possesses this property and will be of great use if mined.
\\
 \hspace*{0.5cm} It has to be mentioned that the conventional methods devised for a single language inevitably fail or give poor results in such cases. Thus to bring more attention of researchers towards this important and challenging aspect, we developed code-mixed corpora for sentiment analysis in Indian languages. India is country with 255 million~\footnote{http://rajbhasha.nic.in/UI/pagecontent.aspx?pc=MzU=} multilingual speakers and one of our goals in this was to challenge the participants and researchers into building advanced and robust systems for sentiment analysis of such code-mixed data. In the present article, we describe the systems and strategies used for making the Bengali-English code-mixed resources. Bengali is an Indo-Aryan language of India where 8.10\% of the total population are the first language speakers and is also the official language of Bangladesh. The original script in which Bengali is written by locals is the Eastern Nagari Script~\footnote{https://www.omniglot.com/writing/bengali.htm}. Majority of our collected data is from Twitter. The reasons why Twitter is an ideal source for collection of such data has been explained by \cite{pak2010twitter}. The contributions of our paper are as follows:
\begin{enumerate}
\item A method for collecting code-mixed data using filtering techniques to assure quality and reduce manual effort.

\item A fast and reliable language identification algorithm (accuracy = 81\%) for code-mixed data with known target languages.

\item A sentiment classification system for code-mixed data using a hybrid system (accuracy = 80.97\%) combining rule based and supervised models.

\item Gold standard Bengali-English code-mixed data with language and polarity tags.

\item Several useful polarity tagged lexicons like phrasal lexicon of length 1200, uni-gram lexicon of length 3000 consisting of phonetically transliterated Bengali words, English acronyms commonly used on social media and a list commonly used emoticons. 
\item Also, a seed list of length 1500 for querying Twitter API for retrieving Bengali-English code-mixed data.
\end{enumerate}

\section{Related Work}
\label{section2}
Several automated systems for Twitter data collection have been made before for corpus collection targeting different aspects but none with the aim to collect code-mixed data as far as our knowledge. On the other hand, various language tagging models have been made recently for code-mixed data and quite a few where a common script has been used for both the languages and one of them is phonetically translated. Among these one of the most relevant works is by \cite{das2014identifying}. Here they demonstrated a system which uses modified character n-gram with weights combined with a lexicon based approach, minimum edit distance as well as context info. \cite{barman2014code} used a hybrid system by combining a lexicon based module with supervised classifiers like SVM, CRF and decision trees. Some of them have also been made as a sub-part for a part-of-speech tagging system like the one by \cite{vyas2014pos}. For sentiment analysis on code-mixed, binary polarity classification has been tried using different classes of supervised models by \cite{ghosh2017sentiment} and for ternary polarity by \cite{ghosh2017complexity} and \cite{sharma2015text}. A comparative study of classifiers trained on different code-mixed features was done by \cite{mandal2018analyzing}. Sophisticated methods using sub-word LSTM for learning sentiments in noisy code-mixed data has been tested as well by \cite{joshi2016towards}.

\section{Code-mixed Corpus Development}
\label{section3}
Corpus collection was done in two steps by collecting raw data from Twitter followed by filtering and cleaning code-mixed data from raw data.

\subsection{Raw Twitter Data Collection}
Our primary aim was to collect quality Bengali-English code-mixed data. However, we observed several instances of phonetically transliterated Bengali utterances (written in Roman script) that do not convey the code-mixed property \cite{muysken2000bilingual}. We were also eager to collect intra-sentential i.e code-switched data instead of inter-sentential since the former is much more common on social media and is relatively more challenging for polarity classification as compared to the latter. For collecting Twitter data, we used the public streaming Twitter API via the Twitter4j~\footnote{http://twitter4j.org/en/} using keywords for querying. The initial keyword list was prepared by considering commonly used positive and negative Bengali words (e.g., bhalo, kharap, baje) and their polarities were validated using Bengali SentiWordNet \cite{das2010sentiwordnet}. We collected a total of 600 code-mixed sentences manually from the initial search output. In order to overcome the saturation problem of the retrieved data with respect to a few query words, we made a validated Bengali keyword list of 1500 unique query words from 600 sentences in decreasing order of frequency. 

\subsection{Data Filtering \& Cleaning}
The collected raw Twitter data contained noise, mostly contributed by words from other languages than the required pair, partially or fully (e.g. \textit{bahar} which is a commonly used Hindi word meaning "outside"), words or full texts not in Roman script, etc. Thus, it was very important to build and apply a filtering module for retaining relatively better quality data in order to reduce manual efforts. Moreover, in order to avoid the problem of duplicacy due to short interval of querying, we have considered two parameters for devising our filtration strategy.
The first parameter is \textalpha\space which denotes the minimum number of Bengali tokens with respect to our seed list whereas \textbeta\space refers to the minimum length of a tweet. It was observed that, the coverage of the top frequent keywords from the seed list helped us to filter majority of our code-mixed instances from the raw data if we vary the values of \textalpha\space only in the range of 1 to 3 and \textbeta\space in between 4 to 6. However, in order to filter more code-mixed instances for fulfilling our requirement, we had to increase the value of \textalpha\space up to 5 and the \textbeta\space up to 8 to maintain the code-mixed property in our filtered tweets. The total amount of raw tweets collected was around 89k and the our filtering system filtered out about 10k tweets from it. The statistics are shown in the Table~\ref{table1}. Here N denotes the information of $n^{\text{th}}$ settings using which the Twitter API was queried, filtered data denotes the number of data remaining after removal.\\
\begin{table}[H]
\centering
\begin{tabular}{|c|c|c|c|c|}
\hline
\textbf{N} & \textbf{\textalpha} & \textbf{\textbeta} & \textbf{Keywords Spent} & \textbf{Filtered Data} \\ \hline
1 & 2 & 4 & 150 & 3800 \\ \hline
2 & 2 & 5 & 250 & 2500 \\ \hline
3 & 3 & 6 & 300 & 1800 \\ \hline
4 & 4 & 7 & 350 & 1500 \\ \hline
5 & 5 & 8 & 450 & 900 \\ \hline
\textbf{sum} & \multicolumn{2}{c|}{} & 1500 & $\approx$ 10500 \\ \hline
\end{tabular}
\caption{Filtering statistics with respect to \textalpha\space and \textbeta\space.}
\label{table1}
\end{table}

During the cleaning process, spams, incomplete tweets, ones with conflicting sentiments were removed manually. Sarcastic tweets were not removed as it has become a very common tool for expression in the 21st century, especially on social media and thus it is important to classify them properly using more advanced techniques. URLs and Hashtags were kept as well as they too are important for sentiment analysis~\footnote{https://open.blockspring.com/bs/sentiment-analysis-from-url-with-alchemyapi} (e.g. visiting the URL for analysis). We wanted to keep the data as untouched as possible to urge the future researchers to build highly robust systems which can be directly used on social media contents without much modification. Table ~\ref{table1.1} show the retrieved, filtered and used code-mixed data counts. It can be seen that our filtering system filtered out quite a lot of data and retained only about 11.79\%.
\\
\begin{table}[H]
\centering
\begin{tabular}{|l|l|l|}
\hline
\textbf{Type}                                                             & \textbf{R} & \textbf{Count} \\ \hline
\textbf{\begin{tabular}[c]{@{}l@{}}Retrieved\\ Tweets (RT)\end{tabular}}  & $\approx$          & 89000          \\ \hline
\textbf{\begin{tabular}[c]{@{}l@{}}Filtered\\ Tweets (FT)\end{tabular}}   & $\approx$          & 10500        \\ \hline
\textbf{\begin{tabular}[c]{@{}l@{}}Code-Mixed\\ Tweets (CT)\end{tabular}} & $\approx$          & 5000     \\ \hline
\end{tabular}
\caption{Tweets retrieved statistics.}
\label{table1.1}
\end{table}
Some examples from our collected data after filtering are given below (underlined - EN, normal - BN) -
\begin{enumerate}
\item Thik \underline{fairy tale} er ending tar moton amra shobaio \underline{happily ever after} thakte lagilam. (Trans: \textit{Just like a fairy tale ending we also lived happily ever after.})
\item \underline{Script} ta khub \underline{tiring} chilo amar mote, aro onek \underline{better} hote parto. (Trans: \textit{The script was very tiring according to me, could have been much better.})
\end{enumerate}
\section{Annotation}
\label{section4} 
In order to annotate the language and sentiment tags to the filtered and cleaned tweets, we developed a system that help in basic annotation. One of the motivations of our annotation task was to reduce the manual tagging effort as we had to deal with huge amount of tweets \textasciitilde 10K. Therefore, in order to cope up with the problems of manual annotations, we planned to build two basic annotation systems, one is for language tagging and another is for sentiment tagging. Both of the annotation systems are described in subsection~\ref{System}. Finally, the outputs of these systems were evaluated by two sets of annotators, one set (A) consisted of a single annotator from Computer Science background with Bengali as mother tongue, where as the second set (B) consisted of five experts and the final evaluation was done by them. In order to handle the confusion cases, an annotation guideline as discussed in subsection ~\ref{section 4.2} was provided to the annotators prior tagging.  

\subsection{System based Annotation}
Out of 10k filtered tweets given by the system, we manually selected a collection  of 5k tweets (as all filtered tweets were not code-mixed) and then we fed it the language tagging and sentiment tagging systems. 
\label{System}
\subsubsection{Language Tagging System}
For language tagging, we used a two-step modular approach by combining lexicon based module (LBM) along with a supervised learning module (SLM). \\ \\
\hspace*{0.25cm}\textbf{LBM:} As our target was simple, that is only to tag Bengali (BN) or English (EN) at word level, we tried to develop a relatively simple system. All  the other unknown words are tagged as UN. The resources used to build the language tagging system are -
\begin{enumerate}[leftmargin=*]
\item A list of Bengali words of size 3000 was prepared from the code-mixed data used in \cite{mandal2018analyzing}. Same words with different phonetic transliterations (e.g. \textit{bhalo} and \textit{balo}) both meaning good were also kept in the list.
\item \textit{English Words (EW)} - A list containing 466k English words~\footnote{https://github.com/dwyl/english-words} was collected from online open sources.
\item \textit{Suffix List (SL)} \& \textit{Acronym List (AL)} - An English suffix list~\footnote{https://www.learnthat.org/pages/view/suffix.html} (e.g. \textit{ing}, \textit{ism}, \textit{ious}) and an English acronym list~\footnote{http://www.muller-godschalk.com/acronyms.html} (e.g. \textit{bbl}-be back later, \textit{omg}- oh my god) was collected.
\item \textit{N-Grams} - Bi-grams and tri-grams dictionary at character level was prepared from the above mentioned Bengali (BW) and English (EW) word lists, where keys were the n-grams and the respective values were frequency.
\end{enumerate}

\hspace*{0.25cm}\textbf{SLM:} A supervised language tagger was developed by training the Linear Support Vector Machine (LSVC) implemented using scikit learn on two features which were character n-grams (n:2,3) as described in LBM features. For training, Bengali word list and list of most common English words~\footnote{http://www.ef.com/english-resources/english- vocabulary/top-3000- words/} were used. The langauge tagging algorithm first searches the target token into our lexicons and if found, the appropriate tag is given. If not found, the supervised tagger is used to output the tag of that target token. The system was tested on ICON 2016~\footnote{http://ltrc.iiit.ac.in/icon2016/} POS tagging contest data and achieved a score of 86.24\%. 
\subsubsection{Sentiment Tagging System}

We used a hybrid system for sentiment classification. Similar to language tagging system, the sentiment tagging system also checks whether a tweet sentence is positive / negative / neutral using rule based method and if it fails, the supervised classifier is employed to produce the output sentiment tag. The resources which were prepared and used in the rule based were also used in supervised method as features.
\\
\\
\hspace*{0.25cm}\textbf{Rule Based Method} - For our rule based checking, three rules that were used to identify the sentiment of a tweet are as follows -
\begin{enumerate}[leftmargin=*]
\item \textit{Feeling (FLNG)} - A regular expression was used to extract the word that follows '$-$ feeling' which is commonly used to express how the author feels. As such instances were self-tagged by the authors, there is no chance of ambiguity with respect to sentiment tagging. These tags are used since the stand alone texts may send different emotional signals or the author might simply be trying to convey his emotions directly.
\item \textit{Hashtag (HT)} - Hashtags which used camel-casing or underscore separation were split and matched with lexicons and n-grams.
\item \textit{Emoticon (EMO)} - Emoticons have a very strong impact on sentence level sentiment. We have used both Unicode and Icon representations of positive and negative emoticons for our experiments. Emoticon scoring has been experimented in three ways, e.g. higher frequency, greater index and average index. The second method which is based on the theory that the emoticon with the greatest index has the greatest influence on the tweet sentiment showed the best results.
\end{enumerate}

\textbf{Supervised Method} - We have experimented with several supervised classifiers. In the Na\"ive Bayes (NB) family, we have used Gaussian Na\"ive Bayes (GNB), Bernoulli Na\"ive Bayes (BNB) and Multinomial Na\"ive Bayes (MNB). The Linear Models (LM) we have tested with are Linear Regression (LRC) and Stochastic Gradient Descent (SGDC). The scikit-learn~\footnote{http://scikit-learn.org/stable/} implementations of the models were used. The features used for supervised methods are as follows -
\begin{enumerate}[leftmargin=*]
\item \textit{Word N-Grams (WN)} - Word level uni-grams, bi-grams and trigrams were adopted as features. Each of the n-grams was sorted according to frequency in non-increasing order and the top 2000 n-grams were selected for training.
\item \textit{Negation (NEGA)} - Negation in a message always reverses its sentiment orientation. If the number of negating words is odd, the polarity is reversed otherwise the calculated polarity is retained. Therefore, we collected a total of 25 English and 130 Bengali unique negation words.
\item \textit{Tagged Words (TGW)} - We also collected 1198 positive  and 1802 negative Bengali uni-grams from an external code-mixed data available in \cite{mandal2018analyzing}. We combined them with English positive and negative words collected from NRC Emotion Lexicon and SOCAL lexicon to build a lexicon containing positive uni-grams (POSU) and negative uni-grams (NEGU).
\item \textit{Tagged Phrases (TGP)} - In addition to words, we made a phrasal lexicon of length 1200 by extracting phrases ($\geq$ 1 from each sentence) from the code-mixed data described in \cite{mandal2018analyzing}. Such phrases are responsible to convey sentiment at the sentence level. For example, "boshe dekha jaye na" (trans: can't sit and watch), "onekei couldn't sleep" (trans: many couldn't sleep), etc. In case of tagged phrases, four scenarios were tested, \textit{perfect match} - the phrase present in the sentence is identical to the tagged phrase, \textit{sparse match} - all the unigrams of the tagged phrase are present in the sentence but not in the same order, \textit{partial match} - a bi-gram from the tagged phrase (if |phrase| $\geq$ 2) is present in the sentence in exact order (a bigram unit of stop-words is not considered) and finally, \textit{no match} - none of the uni-grams is matched or the matched uni-gram is a stop-word.
\item \textit{Tagged Acronyms (TA)} - Commonly used abbreviations on social networking sites were collected and polarity tagged as either positive or negative.
\item \textit{SentiWordNet 3.0 (SWN)} - A word appeared in SentiWordNet \cite{baccianella2010sentiwordnet} containing scores positive, negative and objective.
\item \textit{SOCAL} - This lexicon is used for calculating semantic orientation \cite{taboada2011lexicon}. For utilizing intensifiers of the lexicon, we used the logic that if both the intensifier and word is positive add their score, if both are negative add their scores and negate, if intensifier is positive and word is negative then subtract intensifier score from word score and finally if intensifier is negative and word is positive then add their score.
\item \textit{NRC Emotion Lexicon} - a list of English words and their association with eight basic emotions and sentiment tags \cite{mohammad2013nrc}. In case of our classifier, we only utilized two polarity tags.
\end{enumerate}
\hspace*{0.25cm}For training our supervised classifiers, we used a manually tagged gold-standard dataset containing a total of 1500 training instances, i.e 500 of each polarity, created by merging data from \cite{mandal2018analyzing} and \cite{ghosh2017sentiment}. In case of testing, we used a total of 600 tweets, i.e 200 of each polarity. The data (training and testing) had no data in common in the released versions. However, the features as mentioned for supervised learning were also used to train these classifiers. Different evaluation parameters scored by each of the classifiers are described in Table ~\ref{tab10}. Other than the \textit{accuracy}, the mean value was considered over the three polarities for each of the other parameters. In Table ~\ref{tab10}, we can clearly find that SGDC achieved the best F1-Score with a value of 78.70. Thus, for building our polarity tagger, we finally used the trained model of SGDC. Paramaters (\textit{Param}) were Accuracy (\textit{Acc.}), Precision (\textit{Prec.}), F1- Score (\textit{F1}) and G-Score (\textit{G}).
\begin{table}[H]
\centering
\begin{tabular}{|l|c|c|c|c|c|}
\hline
\multicolumn{1}{|l|}{} & \multicolumn{3}{c|}{\textbf{Na\"ive Bayes (NB)}}      & \multicolumn{2}{c|}{\textbf{Linear Model (LM)}} \\ \hline
\textit{Param}    & \textit{GNB} & \textit{BNB} & \textit{MNB} & \textit{SGDC}       & \textit{LRC}     \\ \hline
Acc.               & 74.83        & 76.16        & 78.16        & 78.66               & 77.00            \\ \hline
Prec.              & 75.05        & 76.25        & 78.56        & 79.20               & 77.40            \\ \hline
Recall                 & 74.83        & 76.16        & 78.16        & 78.66               & 77.00            \\ \hline
F1               & 74.87        & 76.17        & 78.18        & \textbf{78.70}      & 77.02            \\ \hline
G              & 74.90        & 76.19        & 78.27        & 78.81               & 77.11            \\ \hline
\end{tabular}
\caption{Performance of different classifiers.}
\label{tab10}
\end{table}

The confusion matrix of the best performing classifier, that is SGDC, is shown in Table~\ref{table4}. We can see that the classifier is quite stable and not very biased towards a single polarity. The best individual polarity accuracy is for neutral tweets (83\%), which again supports the point regarding it's stability.

\begin{table}[H]
\centering
\begin{tabular}{|c|c|c|c|}
\hline
\textit{}    & \textit{pos} & \textit{neg} & \textit{neu} \\ \hline
pos & 161          & 12           & 27           \\ \hline
neg & 17           & 145          & 38           \\ \hline
neu & 13           & 21           & 166          \\ \hline
\end{tabular}
\captionsetup{justification=centering}
\caption{Confusion matrix of SGDC classifier (italics - predicted values, roman - true values).}
\label{tab11}
\end{table}

The final algorithm we used for sentiment tagging by combining rule based and supervised into a hybrid routine is described below -
\\
\\
Input $\gets$ sentence\\
Output $\to$ polarity
\\
\\
\\
\textbf{Step 1}: \textbf{if} FLNG (sentence) $\neq$ neutral \textbf{then}\\
            \hspace*{1.1cm}\textbf{return} FLNG (sentence) \textbf{else} goto Step 2\\
\textbf{Step 2}: \textbf{if} EMO (sentence) $\neq$ neutral \textbf{then}
            \\\hspace*{1.1cm}\textbf{return} EMO (sentence) \textbf{else} goto Step 3\\
\textbf{Step 3}: \textbf{if} HT (sentence) $\neq$ neutral \textbf{then}
            \\\hspace*{1.1cm}\textbf{return} HT (sentence) \textbf{else} goto Step 4\\
\textbf{Step 4}: \textbf{return} SGDC (sentence)
\\	 
\\
Here FLNG, EMO and HT are the functions described under rule based methods in feature section and SGDC is our trained supervised classifier. 

\subsection{Annotators' Guidelines}
\label{section 4.2}
As the data is already language and sentiment tagged by the systems, the manual annotation efforts were reduced drastically. However, in order to prepare a gold standard corpus with good quality, we finally handed it over to our annotators along with a number guidelines. We provided a very less number of guidelines as most of the urgent issues were already considered by using our systems. \\
\\
\textbf{Language Tagging} - In case of language tagging, the scope of the current target word and the words preceding and succeeding the target word were considered.
\\
\\
\textit{Bengali (BN) \& English (EN) Tag} 
\vspace{-3mm}
\begin{itemize}
\item [\textbf{LG1}] The word is present in the respective language dictionary or is a slang or acronym of that language.\\
e.g. "\textit{hall}" tagged as EN and "\textit{ghor}" tagged as BN.
\item [\textbf{LG2}] whether the word in context belongs to that respective language or not.\\
e.g. "\textit{bar}" in "\textit{onek bar bolechi}" is tagged as BN.  
\item [\textbf{LG3}] The word has any English/Bengali prefix or any English/Bengali suffix.\\
e.g. "\textit{hall is}" tagged to EN and "\textit{ghor ta}" tagged to BN.
\end{itemize}
\textit{Unknown (UN) Tag}
\vspace{-3mm}
\begin{itemize}
\item [\textbf{LG4}] The word does not belong to Bengali or English.\\
e.g. "\textit{amr}" is tagged to UN.
\item [\textbf{LG5}] The token is not recognized (like misspelled words).\\
e.g. "\textit{ankushloveuall}" is tagged to UN.
\item [\textbf{LG6}] The token is a special character, emoticon, URL, etc.\\
e.g. "\textit{@}" is tagged as UN.
\end{itemize}
\textbf{Sentiment Tagging} - For polarity tagging, the authors' perspectives were taken into account and the emotions conveyed from the overall tweet were considered as well.
\\
\\
\textit{Positive Tag \& Negative Tag}
\vspace{-3mm}
\begin{itemize}
\item [\textbf{SG1}] The tweet clearly expresses the sentiment towards the aspect term, for example a person, group or an object.\\
e.g. "\textit{Sir, Boss 2 hit movie hobe. Eid ar sera movie.}" is tagged as positive.
\item [\textbf{SG2}] The tweet clearly expresses the polarity state in mind of the author.\\
e.g. "\textit{Dhurr ar posachhe na all these things.}" is tagged as negative.
\item [\textbf{SG3}] The tweet clearly reports a polar sentiment or mood which may or may not be attributed directly by the author.\\
e.g. "\textit{@username1 yes ami @username2 dadar pagol fan onek diner.}" is tagged as positive.
\end{itemize}
\textit{Neutral Tag}
\vspace{-3mm}
\begin{itemize}
\item [\textbf{SG4}] The tweet contains a mere observation or mention of an objective fact.\\
e.g. "\textit{Dure oi yellow building ta holo shopping mall.}" is tagged as neutral.
\item [\textbf{SG5}] It does not particularly convey any state of mind or opinion. A neutral sentiment is expressed towards the aspect term(s).\\
e.g. "\textit{Cinema ta release koreche."} is tagged as neutral.
\end{itemize}
\textbf{Conflicts} - The confusions occurred during annotation were tabulated as follows
\\
\\
\textit{English (EN) Tag}
\vspace{-3mm}
\begin{enumerate}
\item In the context of a word that contains numerical values were considered by the annotators. For example '11 AM' was tagged as EN by Annotator A while Annotator B tagged "11" as UN and "AM" as EN, separately. 
\item Country names were tagged as EN and UN by Annotator A and B, respectively.
\item Universal words such as, "\textit{table}" were tagged as EN by Annotator A and BN by Annotator B.
\item Words such as "\textit{to}" were tagged as both EN and BN depending on the context and their phonetic representations. 
\end{enumerate} 
\vspace{-1mm}
\textit{Bengali (BN) Tag}
\vspace{-3mm}
\begin{enumerate}
\item The role of the suffix in a word was also dealt ambiguously. For example "\textit{film}" is tagged as EN whereas "\textit{film (ta)}" was tagged as BN. 
\end{enumerate}
\textit{Unknown (UN) Tag}
\vspace{-3mm}
\begin{enumerate}
\item Numerical values such as "1", "2" were tagged as UN. 
\end{enumerate}
We considered two sets of human annotators A and B along with system as the third set. The inter annotator agreement values or Cohen's Kappa (K) are shown in Table~\ref{table4} with respect to each pairs of annotators. In case of sentiment tagging, the annotators agreed on majority of the tweets. However, in both language as well as sentiment tagging, the agreement scores between the sets of manual annotators were comparatively better than the agreements that were calculated with respect to systems. One of the reasons that degrades the system results is relatively small set of training instances in case of both language and sentiment tagging. The annotation details of the system and human annotators are shown in Table~\ref{table5}. 
\begin{table}[H]
\centering
\begin{tabular}{|l|c|}
\hline
\multicolumn{2}{|c|}{\textbf{Language Tagging - Kappa}}  \\ \hline
Annotator A-System                      & 0.69               \\ \hline
Annotator B-System                      & 0.65               \\ \hline
Annotator A-Annotator B                 & 0.83               \\ \hline
\multicolumn{2}{|c|}{\textbf{Sentiment Tagging - Kappa}} \\ \hline
Annotator A-System                      & 0.83               \\ \hline
Annotator B-System                      & 0.82               \\ \hline
Annotator A-Annotator B                 & 0.94               \\ \hline
\end{tabular}
\caption{Inter annotator agreement.}
\label{table4}
\end{table}

\begin{table}[H]
\centering
\begin{tabular}{|c|c|c|c|}
\hline
\multicolumn{4}{|c|}{\textbf{Training Data}}                                                                                                                                  \\ \hline
\multirow{2}{*}{\textbf{}}              & \multicolumn{3}{c|}{\textbf{Language Tag}}                                                                                          \\ \cline{2-4} 
                                        & \textbf{BN Tag}                            & \textbf{EN Tag}                            & \textbf{UN Tag}                           \\ \hline
\textbf{System}                         & 22801                                      & 15130                                      & 331                                       \\ \hline
\textbf{Annotator A}                    & 22460                                      & 15478                                      & 324                                       \\ \hline
\textbf{Annotator B}                    & 22471                                      & 15471                                      & 320                                       \\ \hline
\multicolumn{1}{|l|}{\multirow{2}{*}{}} & \multicolumn{3}{c|}{\textbf{Sentiment Tag}}                                                                                         \\ \cline{2-4} 
\multicolumn{1}{|l|}{}                  & \multicolumn{1}{l|}{\textbf{Pos Tag}} & \multicolumn{1}{l|}{\textbf{Neg Tag}} & \multicolumn{1}{l|}{\textbf{Neu Tag}} \\ \hline
\textbf{System}                         & 988                                        & 926                                        & 586                                       \\ \hline
\textbf{Annotator A}                    & 1010                                       & 987                                        & 503                                       \\ \hline
\textbf{Annotator B}                    & 1000                                       & 1000                                       & 500                                       \\ \hline
\multicolumn{4}{|c|}{\textbf{Testing Data}}                                                                                                                                   \\ \hline
\multicolumn{1}{|l|}{\multirow{2}{*}{}} & \multicolumn{3}{c|}{\textbf{Language Tag}}                                                                                          \\ \cline{2-4} 
\multicolumn{1}{|l|}{}                  & \textbf{BN Tag}                            & \textbf{EN Tag}                            & \textbf{UN Tag}                           \\ \hline
\textbf{System}                         & 22896                                      & 12129                                       & 421                                      \\ \hline
\textbf{Annotator A}                    & 22418                                      & 12620                                       & 408                                      \\ \hline
\textbf{Annotator B}                    & 22416                                      & 12616                                       & 414                                      \\ \hline
\multicolumn{1}{|l|}{\multirow{2}{*}{}} & \multicolumn{3}{c|}{\textbf{Sentiment Tag}}                                                                                         \\ \cline{2-4} 
\multicolumn{1}{|l|}{}                  & \textbf{Pos Tag}                      & \textbf{Neg Tag}                      & \textbf{Neu Tag}                      \\ \hline
\textbf{System}                         & 1077                                       & 642                                        & 741                                       \\ \hline
\textbf{Annotator A}                    & 1094                                       & 698                                        & 668                                       \\ \hline
\textbf{Annotator B}                    & 1090                                       & 705                                        & 665                                       \\ \hline
\end{tabular}
\captionsetup{justification=centering}
\caption{Annotation details of system and human annotators.}
\label{table5}
\end{table}

\section{Corpus Aspect Analysis}
\label{section5}
The released data distribution is shown in Table~\ref{table6}. In both training and testing, the quantity of neutral data is comparatively less as we found that most of the tweets we mined had a polarity. Here, we have analyzed different aspects of our developed gold standard data like code-mixing complexity and generic language aspects. Statistics on some of sentiment affecting aspects like polarity word count, emoticons count, etc were also carried out.
\begin{table}[H]
\centering
\begin{tabular}{|c|c|c|c|}
\hline
\multicolumn{4}{|c|}{\textbf{Distribution}}                                       \\ \hline
\textbf{Purpose}       & \textbf{Positive} & \textbf{Negative} & \textbf{Neutral} \\ \hline
\textbf{Training Data} & 1000              & 1000              & 500             \\ \hline
\textbf{Testing Data}  & 1090              & 705           & 665          \\ \hline
\end{tabular}
\caption{Data distribution.}
\label{table6}
\end{table}

\textbf{Language Aspects}$\hspace{0.1cm}$- Here we analyzed both complexity aspect contributed by code-mixing property (shown in Table~\ref{table7}) as well as other aspects like polarity token counts and mean length (shown in Table~\ref{table8}). Code-Mixing Index (CMI) introduced by \cite{das2014identifying} indicates us the amount of code-mixing found in discourse. Another metric we have calculated which shows the complexity of multilingual corpus is the Complexity Factor (CF) proposed by \cite{ghosh2017sentiment}. CF takes into account three factors- language (LF), switching (SF) and mix (MF) factors. CF was calculated using all the three methods mentioned in that paper. From Table~\ref{table7}, we have observed that the collected code-mixed data has a higher code-mixing index as compared to FIRE 2015~\footnote{http://fire.irsi.res.in/fire/2015/home} Shared Task Corpus (CMI = 11.65) and ICON 2015~\footnote{http://ltrc.iiit.ac.in/icon2015/} Shared Task Corpus (CMI = 5.73). Thus, we can conclude that our data is more complex from code-mixing point of view as compared to FIRE and ICON corpus. We can also see that on an average, positive data has higher code-mixing as compared to other polarities while neutral has comparatively lower code-mixing. From the training and testing values we can also see that the variance is quite nominal, thus adding to the quality of prepared corpus.

\begin{table}[H]
\centering
\begin{tabular}{|l|l|l|l|l|l|l|l|}
\hline
\multicolumn{2}{|c|}{\textbf{}}                                        & \multicolumn{3}{c|}{\textbf{Training}}                                                                    & \multicolumn{3}{c|}{\textbf{Testing}}                                                                     \\ \hline
\multicolumn{1}{|c|}{\textbf{index}} & \multicolumn{1}{c|}{\textbf{f}} & \multicolumn{1}{c|}{\textbf{pos}} & \multicolumn{1}{c|}{\textbf{neg}} & \multicolumn{1}{c|}{\textbf{neu}} & \multicolumn{1}{c|}{\textbf{pos}} & \multicolumn{1}{c|}{\textbf{neg}} & \multicolumn{1}{c|}{\textbf{neu}} \\ \hline
\multirow{3}{*}{\textbf{CMI}}        & min                             & 4.02                              & 4.24                              & 4.20                              & 4.16                              & 4.20                              & 4.18                              \\ \cline{2-8} 
                                     & max                             & 50.0                              & 48.6                              & 46.2                              & 48.6                              & 48.6                              & 47.5                              \\ \cline{2-8} 
                                     & mean                            & 31.0                              & 27.9                              & 22.6                              & 23.4                              & 21.6                              & 20.0                              \\ \hline
\multirow{3}{*}{\textbf{CF1}}        & min                             & 0.38                              & 0.52                              & 0.46                              & 0.44                              & 0.46                              & 0.45                              \\ \cline{2-8} 
                                     & max                             & 20.8                              & 18.0                              & 23.0                              & 37.5                              & 23.0                              & 37.5                              \\ \cline{2-8} 
                                     & mean                            & 4.20                              & 3.93                              & 3.71                              & 4.14                              & 3.67                              & 4.17                              \\ \hline
\multirow{3}{*}{\textbf{CF2}}        & min                             & 4.58                              & 4.81                              & 4.76                              & 4.63                              & 4.76                              & 4.72                              \\ \cline{2-8} 
                                     & max                             & 57.5                              & 62.4                              & 54.8                              & 69.2                              & 64.6                              & 69.2                              \\ \cline{2-8} 
                                     & mean                            & 26.1                              & 24.4                              & 20.5                              & 23.3                              & 21.4                              & 20.7                              \\ \hline
\multirow{3}{*}{\textbf{CF3}}        & min                             & 4.25                              & 4.41                              & 4.36                              & 4.27                              & 4.36                              & 4.31                              \\ \cline{2-8} 
                                     & max                             & 53.8                              & 58.4                              & 51.8                              & 68.0                              & 61.5                              & 68.0                              \\ \cline{2-8} 
                                     & mean                            & 24.2                              & 22.6                              & 19.1                              & 21.6                              & 19.9                              & 19.3                              \\ \hline
\end{tabular}
\caption{Complexity statistics (f - function).}
\label{table7}
\end{table}
Other important language related aspects are are shown in Table~\ref{table8}. The relation for negation count is $\geq$ as lexical checking was done so whereas there might be more number of negations. The aspect values were calculated based on post annotator tagging of language and sentiment. The probable reason for higher negation in negative data is mainly because of the habit of users to express negative sentiment by negating positive words, e.g. \textit{bhalo na} which means "not good". This can be confirmed as well by skimming through the data. The table also tells us that users tend to write relatively more to the point and short tweets while expressing negative sentiments. This is checked from the mean length and UN word count values. Also, BN/EN ratio tells us that users tend to use more Bengali words for expressing objective sentiments.
\par
\begin{table}[H]
\centering
\begin{tabular}{|l|l|l|l|l|l|}
\hline
\multicolumn{6}{|c|}{\textbf{Language Aspects}}                                           \\ \hline
\textbf{}  & \multicolumn{5}{c|}{\textbf{Training Data}}                                       \\ \hline
\textbf{N} & \textbf{Attribute} & \textbf{R} & \textbf{Pos} & \textbf{Neg} & \textbf{Neu} \\ \hline
1          & Negation Count     & $\geq$     & 148          & 449          & 170          \\ \hline
2          & Mean Length        &  =         & 18.50        & 18.06        & 17.91        \\ \hline
3          & BN word count      &   =        & 8541         & 8866         & 5064         \\ \hline
4          & EN word count      &   =        & 6997         & 6535         & 1939         \\ \hline
5		& UN word count &=& 110 &93&117\\ \hline
6          & BN/EN Ratio        &  =         & 1.220        & 1.356        & 2.611        \\ \hline
           & \multicolumn{5}{c|}{\textbf{Testing Data}}                                        \\ \hline
1          & Negation Count     & $\geq$     & 182          & 375          & 200          \\ \hline
2          & Mean Length        &  =         & 18.94        & 16.23        & 17.46        \\ \hline
3          & BN word count      &   =        & 8664        & 7388         & 6364         \\ \hline
4          & EN word count      & =          & 5985         & 4329         & 2302         \\ \hline
5		& UN word count &=& 168 &118&128\\ \hline
6          & BN/EN Ratio        &  =         & 1.447        & 1.706        & 2.764        \\ \hline
\end{tabular}
\caption{Language statistics. (R - relation)}
\label{table8}
\end{table}

\textbf{Emotion Aspects}$\hspace{0.1cm}$- Statistics of sentiment affecting aspects are shown in Table~\ref{table9}. Users tend to explicitly convey their feelings by using the feeling tag more so while expressing negative sentiment as compared to positive. For emoji count the relation is $\geq$ as lexical checking was done, so in reality there might be more number of emoticons. Same is the case for polarity word count, but here $\approx$ is used instead as contextually the word may not be positive or negative. From positive and negative word count in Table~\ref{table9} we can see that users tend to use English polarity words more often as compared to Bengali while expressing.  
\begin{table}[H]
\centering
\begin{tabular}{|l|l|l|l|l|l|}
\hline
\multicolumn{6}{|c|}{\textbf{Sentiment Aspects}}                                                                                                                                                                                                                            \\ \hline
\multicolumn{1}{|c|}{\textbf{}}  & \multicolumn{5}{c|}{\textbf{Training Data}}                                                                                                                                                                                                   \\ \hline
\multicolumn{1}{|c|}{\textbf{N}} & \multicolumn{1}{c|}{\textbf{Attribute}} & \multicolumn{1}{c|}{\textbf{R}} & \multicolumn{1}{c|}{\textbf{Pos}}                   & \multicolumn{1}{c|}{\textbf{Neg}}                  & \multicolumn{1}{c|}{\textbf{Neu}}                 \\ \hline
1                                & POS emoji count                         &  $\geq$                               & 18                                                  & 2                                                  & 2                                                \\ \hline
2                                & NEG emoji count                         &  $\geq$                               & 3                                                   & 17                                                 & 1                                                \\ \hline
3                                & POS word count                          &   $\approx$                              & \begin{tabular}[c]{@{}l@{}}1187/\\ 587\end{tabular} & \begin{tabular}[c]{@{}l@{}}118/\\ 51\end{tabular} & \begin{tabular}[c]{@{}l@{}}35/\\ 26\end{tabular} \\ \hline
4                                & NEG word count                          &  $\approx$                               & \begin{tabular}[c]{@{}l@{}}103/\\ 65\end{tabular}  & \begin{tabular}[c]{@{}l@{}}757/\\ 416\end{tabular} & \begin{tabular}[c]{@{}l@{}}32/\\ 19\end{tabular} \\ \hline
5                                & Feeling tag count                       & =                               & 5                                                   & 10                                                 & 1                                                 \\ \hline
                                 & \multicolumn{5}{c|}{\textbf{Testing Data}}                                                                                                                                                                                                    \\ \hline
1                                & POS emoji count                         & $\geq$                                & 22                                                 & 5                                                 & 3                                               \\ \hline
2                                & NEG emoji count                         & $\geq$                                & 6                                                 & 20                                                & 1                                                \\ \hline
3                                & POS word count                          &  $\approx$                               & \begin{tabular}[c]{@{}l@{}}918/\\ 435\end{tabular}  & \begin{tabular}[c]{@{}l@{}}106/\\ 42\end{tabular} & \begin{tabular}[c]{@{}l@{}}27/\\ 19\end{tabular} \\ \hline
4                                & NEG word count                          &  $\approx$                               & \begin{tabular}[c]{@{}l@{}}119/\\ 72\end{tabular}   & \begin{tabular}[c]{@{}l@{}}673/\\ 341\end{tabular} & \begin{tabular}[c]{@{}l@{}}28/\\ 13\end{tabular} \\ \hline
5                                & Feeling tag count                       & =                               & 4                                                   & 8                                                  & 2                                                 \\ \hline
\end{tabular}
\captionsetup{justification=centering}
\caption{Sentiment affecting aspects. For POS, NEG word count representation format is EN/BN. (R - relation)}
\label{table9}
\end{table}
\textbf{Other Aspects}$\hspace{0.1cm}$-
The most common polarity carrying words from the code-mixed data are shown in Table~\ref{table10}. From the table we can see that the most common polar words are highly polar. These words are commonly used while speaking as well. It can also be seen that a lot of counterparts are present in the table, like bhalo - good, osadharon - special, kharap - bad, betha - pain, etc. 
\begin{table}[H]
\centering
\begin{tabular}{|l|l|l|}
\hline
\multicolumn{3}{|c|}{\textbf{Most Common Words (freq>150)}}                                                                                                                                                                                 \\ \hline
\textbf{}         & \multicolumn{1}{c|}{\textbf{Bengali}}                                                                 & \multicolumn{1}{c|}{\textbf{English}}                                                                \\ \hline
\textbf{Positive} & \begin{tabular}[c]{@{}l@{}}bhalo, besh,\\ shundor, darun,\\ moja, pochondo,\\ osadharon\end{tabular}   & \begin{tabular}[c]{@{}l@{}}love, best,\\ good, comedy,\\ better, special,\\ famous, happy\end{tabular} \\ \hline
\textbf{Negative} & \begin{tabular}[c]{@{}l@{}}kharap, baje,\\ kosto, boka,\\ bekar, chinta,\\ jhogra, betha\end{tabular} & \begin{tabular}[c]{@{}l@{}}poor, bad,\\ problem, old,\\ sad, busy,\\ bogus, pain\end{tabular}        \\ \hline
\end{tabular}
\captionsetup{justification=centering}
\caption{Some common Bengali and English words, training and testing data combined.}
\label{table10}
\end{table}

\section{System Performance on Final Data}
\label{section7}
After the final annotation was done we tested our systems again on the new gold-standard data. Both the language tagging system and sentiment tagging system (SGDC) was trained on the training data and evaluated on the testing data. The language tagger performed surprisingly well and got an accuracy of 81\%. With the sentiment tagging system we expected a significant improvement due the increased size of the training data. It indeed performed better and got an accuracy of 80.97\% and F1-Score of 81.2\%. In future we would like to test different feature combinations and add contextual features as well to improve our system.    

\section{Release Format}
\label{section8}
The final gold-standard dataset is available in JSON format. We have chosen JSON since it is more compact, lightweight, flexible and easier to use compared to XML. CSV was ignored as well since we needed to represent a hierarchical structure which is much easier with JSON as well. Another problem with CSV is that a standard reader application (e.g. Excel) is quite slow at opening large files as well as unstructured encoded values and spilling. The objects/values provided in the released JSON file are id (data number), lang\textunderscore tagged\textunderscore text (language tagged text), sentiment (-1 $\leftarrow$ negative, 0 $\leftarrow$ neutral, 1 $\leftarrow$ positive) and text (without language tag). A single sample from the JSON file is given below - \\ \\
\hspace*{0.5cm}id: 83 \\
\hspace*{0.5cm}lang\textunderscore tagged\textunderscore text: Onekdin\textbackslash bn por\textbackslash bn spotlight\textbackslash en e\textbackslash bn \hspace*{0.5cm}fire\textbackslash bn eshe\textbackslash bn nijeke\textbackslash bn besh\textbackslash bn bikheto\textbackslash bn bikheto\textbackslash bn \hspace*{0.5cm}lagche\textbackslash bn ,\textbackslash un I\textbackslash en am\textbackslash en toh\textbackslash bn very\textbackslash en hpy\textbackslash en .\textbackslash un \\
\hspace*{0.5cm}sentiment: 1 \\
\hspace*{0.5cm}text: Onekdin por spotlight e fire eshe nijeke besh \hspace*{0.5cm}bikheto bikheto lagche, I am toh very happy.

\section{Conclusion \& Future Work}
\label{section9}
In this paper we have described the steps involved in building the system which we have used for collecting and preparing gold-standard Bengali-English code-mixed data for sentiment analysis. To the best of our knowledge, it is the first publicly released data of its kind. The data we present also has a reliable inter-annotator agreement, K - 0.83 for language tag and K - 0.94 for sentiment tag. We also discuss the challenges faced in each step which should be overcome in future for an improved system. In future, we wish to improve the quality of our system by increasing the population size of our resources and training our classifiers on bigger data. We also wish to find a correlation between \textalpha\space (BN token count) and the keyword used for querying to the API so that the value of \textalpha\space can be varied automatically using computationally calculated rules to fetch more relevant data which in this case was Bengali-English code-mixed.

\section*{References}
\label{main:ref}

\bibliographystyle{lrec}
\bibliography{xample}

\begin{thebibliography}{}

\bibitem[\protect\citename{Baccianella \bgroup et al.\egroup
  }2010]{baccianella2010sentiwordnet}
Baccianella, S., Esuli, A., and Sebastiani, F.
\newblock (2010).
\newblock Sentiwordnet 3.0: An enhanced lexical resource for sentiment analysis
  and opinion mining.
\newblock In {\em LREC}, volume~10, pages 2200--2204.

\bibitem[\protect\citename{Barman \bgroup et al.\egroup }2014]{barman2014code}
Barman, U., Das, A., Wagner, J., and Foster, J.
\newblock (2014).
\newblock Code mixing: A challenge for language identification in the language
  of social media.
\newblock In {\em Proceedings of The First Workshop on Computational Approaches
  to Code Switching}, pages 13--23.

\bibitem[\protect\citename{Das and Bandyopadhyay}2010]{das2010sentiwordnet}
Das, A. and Bandyopadhyay, S.
\newblock (2010).
\newblock Sentiwordnet for bangla.
\newblock {\em Knowledge Sharing Event-4: Task}, 2.

\bibitem[\protect\citename{Das and Gamb{\"a}ck}2014]{das2014identifying}
Das, A. and Gamb{\"a}ck, B.
\newblock (2014).
\newblock Identifying languages at the word level in code-mixed indian social
  media text.

\bibitem[\protect\citename{Ghosh \bgroup et al.\egroup
  }2017a]{ghosh2017complexity}
Ghosh, S., Ghosh, S., and Das, D.
\newblock (2017a).
\newblock Complexity metric for code-mixed social media text.
\newblock {\em arXiv preprint arXiv:1707.01183}.

\bibitem[\protect\citename{Ghosh \bgroup et al.\egroup
  }2017b]{ghosh2017sentiment}
Ghosh, S., Ghosh, S., and Das, D.
\newblock (2017b).
\newblock Sentiment identification in code-mixed social media text.
\newblock {\em arXiv preprint arXiv:1707.01184}.

\bibitem[\protect\citename{Joshi \bgroup et al.\egroup }2016]{joshi2016towards}
Joshi, A., Prabhu, A., Shrivastava, M., and Varma, V.
\newblock (2016).
\newblock Towards sub-word level compositions for sentiment analysis of
  hindi-english code mixed text.
\newblock In {\em COLING}, pages 2482--2491.

\bibitem[\protect\citename{Mandal and Das}2018]{mandal2018analyzing}
Mandal, S. and Das, D.
\newblock (2018).
\newblock Analyzing roles of classifiers and code-mixed factors for sentiment
  identification.
\newblock {\em arXiv preprint arXiv:1801.02581}.

\bibitem[\protect\citename{Mohammad and Turney}2013]{mohammad2013nrc}
Mohammad, S.~M. and Turney, P.~D.
\newblock (2013).
\newblock Nrc emotion lexicon.
\newblock Technical report, NRC Technical Report.

\bibitem[\protect\citename{Muysken}2000]{muysken2000bilingual}
Muysken, P.
\newblock (2000).
\newblock {\em Bilingual speech: A typology of code-mixing}, volume~11.
\newblock Cambridge University Press.

\bibitem[\protect\citename{Pak and Paroubek}2010]{pak2010twitter}
Pak, A. and Paroubek, P.
\newblock (2010).
\newblock Twitter as a corpus for sentiment analysis and opinion mining.
\newblock In {\em LREc}, volume~10.

\bibitem[\protect\citename{Sharma \bgroup et al.\egroup }2015]{sharma2015text}
Sharma, S., Srinivas, P., and Balabantaray, R.~C.
\newblock (2015).
\newblock Text normalization of code mix and sentiment analysis.
\newblock In {\em Advances in Computing, Communications and Informatics
  (ICACCI), 2015 International Conference on}, pages 1468--1473. IEEE.

\bibitem[\protect\citename{Taboada \bgroup et al.\egroup
  }2011]{taboada2011lexicon}
Taboada, M., Brooke, J., Tofiloski, M., Voll, K., and Stede, M.
\newblock (2011).
\newblock Lexicon-based methods for sentiment analysis.
\newblock {\em Computational linguistics}, 37(2):267--307.

\bibitem[\protect\citename{Vyas \bgroup et al.\egroup }2014]{vyas2014pos}
Vyas, Y., Gella, S., Sharma, J., Bali, K., and Choudhury, M.
\newblock (2014).
\newblock Pos tagging of english-hindi code-mixed social media content.
\newblock In {\em EMNLP}, volume~14, pages 974--979.

\end{thebibliography}

\end{document}